\documentclass[sigconf]{acmart}

\AtBeginDocument{%
  \providecommand\BibTeX{{%
    \normalfont B\kern-0.5em{\scshape i\kern-0.25em b}\kern-0.8em\TeX}}}

\copyrightyear{2023}
\acmYear{2023}
\setcopyright{acmlicensed}
\acmConference[MM '23] {Proceedings of the 31st ACM International Conference on Multimedia}{October 29--November 3, 2023}{Ottawa, ON, Canada.}
\acmBooktitle{Proceedings of the 31st ACM International Conference on Multimedia (MM '23), October 29--November 3, 2023, Ottawa, ON, Canada}
\acmPrice{15.00}
\acmISBN{979-8-4007-0108-5/23/10}
\acmDOI{10.1145/3581783.3613845}

\usepackage{microtype}
\usepackage{booktabs}
\usepackage{color,colortbl}

\usepackage{amsmath}
\usepackage[bb=dsserif]{mathalpha}
\usepackage{esvect}

\usepackage{caption}
\usepackage{subcaption}
\usepackage{graphicx}
\usepackage{float}
\usepackage{bbding}

\usepackage{multirow}

\newcommand{\Lagr}{\mathcal{L}}

\newcommand{\W}{\mathcal{W}}

\DeclareMathOperator*{\argmax}{arg\,max}

\begin{document}

\title{Towards Visual Taxonomy Expansion}

\author{Tinghui Zhu}
\authornote{Work done when interned at Meituan.}
\email{thzhu22@m.fudan.edu.cn}
\affiliation{%
  \institution{Shanghai Key Laboratory of Data Science, School of Computer Science, Fudan University}
  \city{Shanghai}
  \country{China}
}

\author{Jingping Liu}
\authornote{Jingping Liu and Yanghua Xiao are the corresponding authors.}
\email{jingpingliu@ecust.edu.cn}
\affiliation{%
  \institution{School of Information Science and Engineering, East China University of Science and Technology}
  \city{Shanghai}
  \country{China}
}

\author{Jiaqing Liang}
\email{liangjiaqing@fudan.edu.cn}
\affiliation{%
  \institution{Shanghai Key Laboratory of Data
Science, School of Computer Science,
Fudan University}
  \city{Shanghai}
  \country{China}
}

\author{Haiyun Jiang}
\email{16110240020@fudan.edu.cn}
\affiliation{%
  \institution{Shanghai Key Laboratory of Data Science, School of Computer Science, Fudan University}
  \city{Shanghai}
  \country{China}
}

\author{Yanghua Xiao}
\authornotemark[2]
\email{shawyh@fudan.edu.cn}
\affiliation{%
  \institution{Shanghai Key Laboratory of Data
Science, School of Computer Science,
Fudan University}
  \city{Shanghai}
  \country{China}
}

\author{Zongyu Wang}
\email{wangzongyu02@meituan.com}
\affiliation{%
  \institution{Meituan}
  \city{Shanghai}
  \country{China}
}

\author{Rui Xie}
\email{rui.xie@meituan.com}
\affiliation{%
  \institution{Meituan}
  \city{Shanghai}
  \country{China}
}

\author{Yunsen Xian}
\email{xianyunsen@meituan.com}
\affiliation{%
  \institution{Meituan}
  \city{Shanghai}
  \country{China}
}
\renewcommand{\shortauthors}{Tinghui Zhu et al.}


\begin{abstract}
Taxonomy expansion task is essential in organizing the ever-increasing volume of new concepts into existing taxonomies.
Most existing methods focus exclusively on using textual semantics, leading to an inability to generalize to unseen terms and the "Prototypical Hypernym Problem."
In this paper, we propose Visual Taxonomy Expansion (VTE), introducing visual features into the taxonomy expansion task.
We propose a textual hypernymy learning task and a visual prototype learning task to cluster textual and visual semantics.
In addition to the tasks on respective modalities, we introduce a hyper-proto constraint that integrates textual and visual semantics to produce fine-grained visual semantics.
Our method is evaluated on two datasets, where we obtain compelling results.
Specifically, on the Chinese taxonomy dataset, our method significantly improves accuracy by 8.75 \%.
Additionally, our approach performs better than ChatGPT on the Chinese taxonomy dataset.
\end{abstract}





\pagenumbering{gobble}
\maketitle
\section{Introduction}



Taxonomies play a crucial role in interpreting semantics and providing valuable knowledge to machines.
In the age of large language models, taxonomies are still essential, especially when dealing with domain concepts.
Among a a wide range of taxonomies available on the Web \citep{miller1995wordnet, bollacker2008freebase, lipscomb2000medical}, product taxonomies have been proven particularly useful in e-commerce applications such as query understanding, recommendation systems, and product labeling.
However, maintaining a taxonomy manually is a laborious task.
Therefore, the taxonomy expansion task has become increasingly valuable.

Existing literature typically emphasizes textual semantics, overlooking the significance of visual semantics.
These approaches involve representing term features and aligning them with parent terms, wherein a well-formulated term representation can streamline the matching process.
Term representations are classified into two main paradigms:
1) Extracting textual semantics from extensive corpora, by employing implicit relational semantics exclusively \citep{DBLP:conf/www/ManzoorLSL20} and by relying solely on corpus data to generate a constrained seed-guided taxonomy \citep{10.1145/3447548.3467308}.
2) Deriving structural textual semantics from existing taxonomies, using various techniques such as local egonet 
 \citep{10.1145/3366423.3380132}, parent-query-child triplet \citep{DBLP:conf/aaai/ZhangSZCSM021}, and mini-paths \citep{10.1145/3394486.3403145}.
The above mentioned methods disregards visual semantics, leading to two issues:
1) Models struggle to understand the semantics of unseen terms, impeding their integration into the taxonomy.
2) The "Prototypical Hypernym Problem" \citep{levy-etal-2015-supervised} occurs in the supervised settings.
This problem arises when a model learns that \texttt{Fruit} is a hypernym of \texttt{Apple} and that \texttt{Apple Juice} is semantically akin to \texttt{Apple}, subsequently making the faulty assumption that \texttt{Fruit} is the hypernym of \texttt{Apple Juice}.
This highlights the importance of expanding taxonomies to encompass vital features beyond textual semantics and lexical relations.
\begin{figure}[t!]
    \centering
    \includegraphics[width=\linewidth]{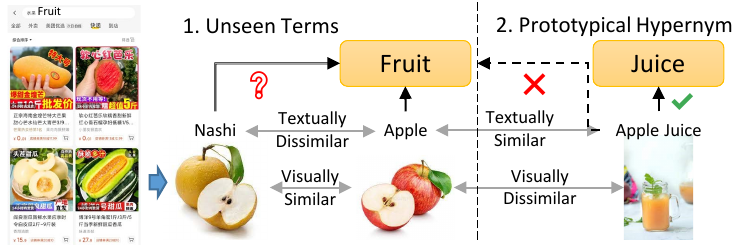}
    \caption{Examples of data sources and how visual feature improves performance on unseen terms and addresses ``Prototypical Hypernym Problem''.}
    \label{fig:intro}
    \vspace{-1em}
\end{figure}
This paper presents Visual Taxonomy Expansion (VTE), a method that enriches term representations by integrating visual semantics derived from images.
Visual semantics help to address the limitations mentioned above.
1) VTE enhances model generalization for the comprehension of unseen terms by using similar visual features.
Specifically, terms with analogous images, such as \texttt{Apple} and \texttt{Nashi}, are more likely to possess a shared hypernym like \texttt{Fruit}.
2) VTE refines the differentiation between prototypical hypernyms and actual hypernyms by distinct visual semantics.
For example, the terms \texttt{Apple} and \texttt{Apple Juice}, which are visually distinct, are less likely to share the same hypernym (\texttt{Fruit}), thereby allowing for differentiation between prototypical hypernyms and real hypernyms (\texttt{Juice}).
However, the incorporation of visual semantics introduces a few challenges:
\begin{itemize}
    \item Coarsely-grained terms, which represent general concepts of several sub-classes, often lack precise images for depiction, making it difficult to find suitable visual semantics.
    \item Even when visual semantics for coarsely-grained terms can be found, establishing connections with corresponding textual semantics becomes challenging, since both semantics represent the same term.
\end{itemize}

To tackle challenges in VTE, we utilize potential hypernymy relations from user-click logs \citep{9835349} and propose a contrastive multitask framework consisting of three representation learning tasks and a detection task.
The first task, textual hypernymy learning, aids in comprehending the relationships between hyponyms and hypernyms by clustering hyponyms to produce superior hypernym representations.
The second task, visual prototype learning, generates visual semantics of hypernyms by grouping similar images using instance-cluster contrastive learning.
The third task, hyper-proto constraint, narrows the gap between textual and visual semantics, resulting in more refined visual prototypes.
It models a variant similarity constrained by uncertainty during projection. Additionally, we develop a heuristic fusion method that incorporates both textual and visual semantics to identify hypernymy relationships using an MLP classifier.

To summarize, our major contributions include the following:
\begin{enumerate}
    \item To the best of our knowledge, we are the first to propose Visual Taxonomy Expansion (VTE). To achieve this goal, we introduce a contrastive multitask framework that generates missing visual features for hypernyms and organically integrates textual and visual semantics.
    \item We provide two datasets, one in Chinese and the other in English, to support the VTE task. The Chinese taxonomy dataset is the largest Chinese visual taxonomy dataset, containing over 10 thousand edges.
    \item We conduct extensive experiments on two taxonomy expansion datasets and achieve the best results over all baselines. Specifically, our method improve the accuracy by 8.75\% on the Chinese taxonomy dataset and 6.91\% on the Semeval-2016 dataset. Moreover, we outperform ChatGPT on the Chinese taxonomy dataset. Our datasets and codes are publicly available\footnote{https://github.com/DarthZhu/VTE}.
\end{enumerate}
\section{Related Work}


\noindent\textbf{Taxonomy Expansion.}
Most previous works on taxonomy expansion have focused on designing features to capture diverse representations of terms.
\citet{DBLP:conf/www/ManzoorLSL20} expand taxonomies by jointly learning latent representations for edge semantics and taxonomy concepts.
\citet{10.1145/3366423.3380132} propose a self-supervised method using a graph neural network to encode positional information for taxonomy expansion.
\citet{10.1145/3447548.3467308} collect information from complex local-structure information and learn to generate concept’s full name from corpus.
\citet{song2021first} design a concept sorting model to extract hyponymy relations and sort their insertion order based on the relationship between newly mined concepts.
Similarly, \citet{10.1145/3077136.3080732} utilize the hierarchical information of the existing taxonomy by extracting tree-exclusive features in the taxonomy to improve the coherence of the resulting taxonomy.
More recently, \citet{jiang2022taxoenrich} leverage both semantic and structural features to generate taxonomy-contextualized embeddings for each term.
\citet{9835349} exploit user-generated content by constructing a user-click graph to augment structural information and used user-click logs as prior hypernymy sources.

\noindent\textbf{Taxonomy Construction.}
For taxonomy construction from scartch, existing methods can be divided into two paradigms.
Pattern-based methods adopt lexical patterns to extract hypernymy relations between co-occurrences of given pairs \citep{hearst-1992-automatic, 10.1145/336597.336644, nakashole-etal-2012-patty, panchenko-etal-2016-taxi}.
Distributional methods predict hypernymy relations by calculating pairwise scores based on term embeddings \citep{shwartz-etal-2016-improving, luu-etal-2016-learning, cocos-etal-2018-comparing, jin2018junction, roller2014inclusive}.
Additionally, some methods employ advanced techniques, such as transfer learning \citep{shang-etal-2020-taxonomy}, reinforcement learning \citep{mao-etal-2018-end}, and entity set expansion techniques \citep{shen2019setexpan}, to construct a taxonomy.


\begin{figure*}[t!]
    \centering
    \includegraphics[width=.95\linewidth]{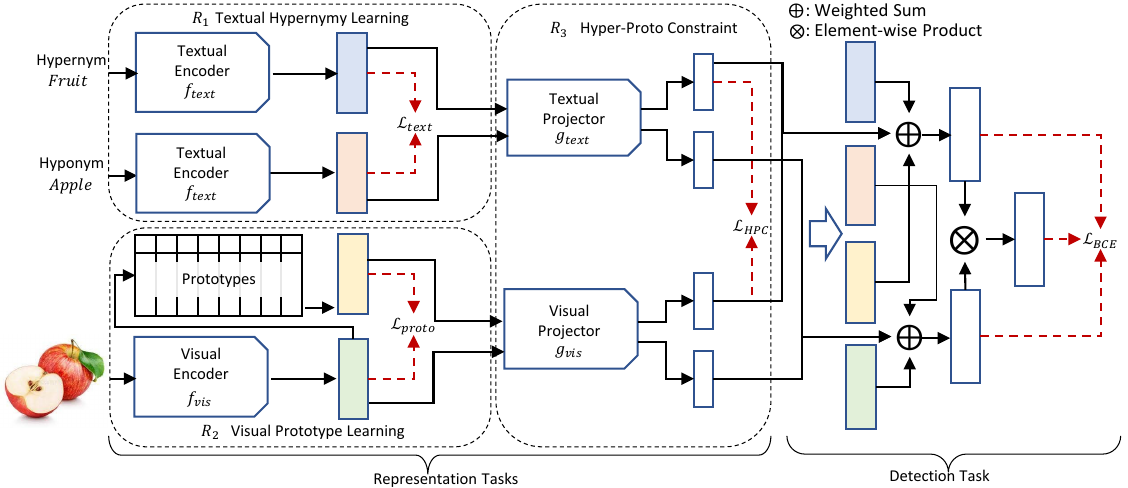}
    \caption{An overview of our proposed framework. Colored representations are the same representations respectively.}
    \label{fig:overview}
\end{figure*}

\section{Overview}
In this section, we firstly define the hypernymy detection task, followed by the framework of our solution.

\subsection{Preliminaries}
\textbf{Definition 1. Taxonomy.} A taxonomy $\mathcal{T}=(\mathcal{N}, \mathcal{E})$ is a tree-structured hierarchy in which each node $n \in \mathcal{N}$ is a term, and each edge $(n^{e}, n^{o}) \in \mathcal{E}$ is a pair of hypernymy relations, indicating that $n^e$ is the hypernym of $n^o$.

\noindent \textbf{Definition 2. User Click Logs.} User click logs $\mathcal{U} = \{(q, I_q)\}$ record a user query $q \in \mathcal{Q}$ and its corresponding clicked item $I_q \in \mathcal{I}$. Each $I_q \in \mathcal{I}$ includes an item name and its image.

\subsection{Problem Formulation}
\label{sec:formulation}
\textbf{Problem Definition.} Given an existing taxonomy $\mathcal{T}=(\mathcal{N}, \mathcal{E})$ and a set of terms $\mathcal{V}$, where for each $v \in V$, $v \not\in N$, the goal of the taxonomy expansion task is to attach each term $v \in V$ to the existing $\mathcal{T}$, such that $\mathcal{T}$ is expanded to $\mathcal{T}^*=(\mathcal{N} \cup \mathcal{V}, \mathcal{E} \cup \mathcal{R})$. Each edge $r \in \mathcal{R}$ represents a new hypernymy pair $r=(n, v)$, where $ n\in \mathcal{N}$ and $v \in \mathcal{V}$.

Formally, we define each node $n \in \mathcal{N}$ as a categorical random variable and the taxonomy $\mathcal{T}$ as a Bayesian network.
The probability of a taxonomy $\mathcal{T}$ is defined as the joint probability of node set $\mathcal{N}$ which can be further decomposed into a product of conditional probabilities as follows:
\begin{equation}
    \mathbf{P}(\mathcal{T}|\theta) = \mathbf{P}(\mathcal{N}|\mathcal{T}; \theta) = \prod_{i=1}^{|\mathcal{N}|}\mathbf{P}(n_i|parent_{\mathcal{T}}(n_i); \theta),
\end{equation}
where $\theta$ is the model parameters and $parent_\mathcal{T}(n_i)$ is the set of parent(s) of the node $n_i$. Then, given an existing taxonomy $\mathcal{T}=(\mathcal{N}, \mathcal{E})$, the taxonomy expansion task is to find the optimal taxonomy $\mathcal{T}^*$ by solving the following optimization problem:
\begin{equation}
    \mathcal{T}^*= \argmax_{\mathcal{T}}\mathbf{P}(\mathcal{T}|\theta) = \argmax_{\mathcal{T}}\sum_{i=1}^{|\mathcal{N} \cup \mathcal{V}|}\mathbf{P}(n_i|parent_\mathcal{T}(n_i); \theta).
\end{equation}

\noindent \textbf{Problem Simplification.} For a given term $v \in \mathcal{V}$, taxonomy expansion tasks find a subset $\mathcal{N}_v \subseteq \mathcal{N} \cup \mathcal{V}$ of parents.
Since all terms $n \in \mathcal{N}$ can be considered as an candidate, leading to an enormous potential search space of ${|\mathcal{N} \cup \mathcal{V}|}$. 
Previous works alleviate this problem via restricting that only one term $n \in \mathcal{N}$ to be the parent of $v \in \mathcal{V}$.
However, this assumption over-simplifies the problem, neglecting the fact that a term often has multiple hypernyms and the connections between new terms in $\mathcal{V}$. 
To further address this problem, we adopt the idea of utilizing strong prior hypernymy relations from user behaviors \citep{9835349}. Potential hypernymy pairs from user click logs $\mathcal{U}$ are treated as the source of candidate parents since when users search for an item, it is more likely that they query a super-class of the item.
Our statistical analysis of 100 query-click pairs yields 69 pairs containing potential hypernymy relations, thereby highlighting $\mathcal{U}$ as an excellent source for obtaining prior hypernymy pairs.


\subsection{Framework}
\label{sec:framework}

As shown in Figure \ref{fig:overview}, our solution framework mainly consists of two types of tasks: representation tasks and detection task.
The former is composed of three sub-tasks: Textual Hypernymy Learning ($R_1$), Visual Prototype Learning ($R_2$), and Hyper-Proto Constraint ($R_3$).

\textbf{Representation tasks.}
$\mathbf{R_1}$: The first task learns the representations of hypernyms and hyponyms in the textual modality. To generate better hypernym representations, hyponyms with the same hypernym are clustered.
$\mathbf{R_2}$: The second task learns the representations of hyponyms and the corresponding prototypes in the visual modality. Identical images are clustered and yield a prototype, which is the cluster centroid of identical images, to represent high-level visual semantics for hyponym images.
$\mathbf{R_3}$: To bridge the gap between the textual and visual representations, the third task learns the similarity between the textual hypernym representations and the visual prototypes. Both representations are projected and aligned in a shared latent space through contrastive learning.



\textbf{Detection task}
is made up of two steps.
The first step takes both textual and visual representations of terms as the input and adopts a heuristic fusion method to generate term representations. The second step detects whether hypernymy relation holds according to these representations.

\section{Methodology}







In this section, we will describe the details of four tasks in our proposed framework.

\subsection{$\boldsymbol{R_1}$: Textual Hypernymy Learning}
\label{sec:text loss}

Textual hypernymy learning learns the representations of hypernyms and hyponyms in the textual modality.
To this end, we adopt contrastive learning to cluster textual-semantically similar terms. In this task, a hyponym and its correct parent are defined as the anchor term and the positive sample, while terms that are not ancestors of the hyponym are treated as negative samples.



Formally, given a pair of terms $X = (x_{hyper}, x_{hypo})$, we define $x_{hypo}$ as the anchor and $x_{hyper}$ as its positive sample while negative samples $x_{neg}^i, 1 \leq i \leq N-1 $, where $N$ is the batch size, are hypernyms of other term pairs in one batch.
To extract informative representations for contrastive learning, we employ a pre-trained language model, e.g., BERT \citep{devlin2019bert}.
Specifically, given a term $x=[w_1, w_2, ..., w_{|x|}]$, we feed $x$ into a pre-trained contextual language model with an additional linear layer $f_{text}$ and obtain the contextual representations $\boldsymbol{S}$ from the last hidden layer:
\begin{equation}
    \boldsymbol{S}=[s_1, s_2, ..., s_{|x|}]=f_{text}([w_1, w_2, ..., w_{|x|}]).
\end{equation}
Finally, we apply mean pooling and a linear transformation on $\boldsymbol{S}$ to generate term-level representation $t = mean(\boldsymbol{S}).$
Such that, for a given pair of terms $X = (x_{hyper}, x_{hypo})$, the encoded textual representations can be denoted as $T = (t_{e}, t_{o})$. Based on the textual representations of given term pairs, we have the textual hypernymy learning objective:
\begin{equation}
    \label{eq:text loss}
    \Lagr_{text}=-\log \frac{{e^{(t_{o}\cdot t_{e}) / \tau_{text}}}}{e^{(t_{o}\cdot t_{e}) / \tau_{text} + \sum_{i=1}^{N-1}{e^{(t_{o}\cdot t_{neg}^i) / \tau_{text}}}}},
\end{equation}
where $\cdot$ is the dot product and $\tau_{text}$ is the temperature hyper-parameter in contrastive learning.

\subsection{$\boldsymbol{R_2}$: Visual Prototype Learning}
\label{sec:dpl}




The visual prototype learning task clusters hyponym images and produces visual prototypes. Visual prototypes are crucial because high-level terms often lack precise images, which leads to a lack of visual semantics. This issue is resolved by adopting prototypical contrastive learning to cluster identical images, which treats an image as an anchor, while the closest prototype is the positive sample, and other prototypes are negative samples.
To represent visual prototypes, vector quantization (VQ) is used, which defines prototypes as a codebook.


Specifically, given a hyponym image $i$, we use a ResNet \citep{he2015deep} encoder with an additional linear layer $f_{vis}$ to encode $i$ to get its visual features $v = f_{vis}(i).$
We assign $v$ to a prototype cluster by defining prototypes as an auxiliary embedding table $\mathcal{P}=\{P_1,P_2,...,P_k\}$ of size $k$.
Each $P_i \in \mathcal{P}$ is a vector of size $e$, representing a prototype vector for a potential cluster.
We calculate the Euclidean distances between $v$ and each $P_i \in \mathcal{P}$, where the closest prototype is chosen.
We can derive the prototype $p=P_i$ for $v$ from the following equation:
\begin{equation}
    \label{eq:argmin}
    i = \mathop{\arg\min}_{j \in \{1, 2, ..., k\}}(||P_j - v||).
\end{equation}
It is noteworthy that the embedding table $\mathcal{P}$ should not be updated by gradients; otherwise, the rapid gradient update will break the consistency with the following cluster selection process. Thus, stop gradient $sg$ is employed on $p$, which is: $p = v + sg(p - v),$ where $sg$ treats the input as a constant during back propagation.

Finally, We employ instance-cluster level contrastive learning to cluster identical images, which prevents degeneration in the optimization process. We denote $v$ as the anchor and the assigned prototype $p$ as the positive sample, and negative samples $p_{neg}^i$, $1\leq i \leq N-1$ denote prototypes chosen by other hyponym images in a batch.
We normalize both $v$ and $p$ and Euclidean distance becomes $(x-y)^2=x^2+y^2-2xy=2-2xy$, which enables us to adopt the InfoNCE loss to simplify the loss computation, where the objective can be written as:
\begin{equation}
    \label{eq:proto}
    \Lagr_{proto}=-\log \frac{{e^{v\cdot p / \tau_{proto}}}}{e^{v\cdot p / \tau_{proto}}+\sum_{i=1}^{N-1}{e^{v\cdot p_{neg}^i / \tau_{proto}}}},
\end{equation}
where $\cdot$ is the dot product and $\tau_{proto}$ is the temperature hyper-parameter in contrastive learning.

To ensure consistent prototypes, we employ Exponential Moving Average (EMA) to optimize $\mathcal{P}$. We update each prototype by images that select it as follows:
\begin{equation}
    \label{eq:ema}
    P_i' = \alpha \times P_i + (1-\alpha)\times\frac{\sum_{j=1}^N{\mathbbb{1}_{p_j=P_i} \times v_j}}{\sum_{j=1}^N{\mathbbb{1}_{p_j=P_i}} + \epsilon},
\end{equation}
where $\epsilon$ is a small positive number that prevents division by zero, and $\alpha$ is the constant smoothing hyper-parameter. We perform momentum update simultaneously with gradient update, ensuring consistency between $f_{vis}$ and $\mathcal{P}$.


\subsection{$\boldsymbol{R_3}$: Hyper-Proto Constraint}
\label{sec:hpc}



The Hyper-Proto Constraint aims to capture the similarity between hypernyms and prototypes. There are two primary reasons why we propose this constraint. Firstly, the visual prototype learning technique we discuss in Section \ref{sec:dpl} clusters images based solely on their visual similarity, which can result in large and noisy clusters. Secondly, we observe that the disparity between textual and visual semantics is substantial. However, quantifying this similarity is challenging because the relationships between hypernyms and prototypes can be diverse. For example, a hypernym such as \texttt{Food} may have many identical visual prototypes, such as those indicating \texttt{Fruit}, \texttt{Vegetables}, and \texttt{Meat}, and vice versa.
To address the challenge, we introduce uncertainty \citep{https://doi.org/10.48550/arxiv.2110.04403}.



Namely, given an encoded hypernym representation $h$ and a corresponding visual prototype $p$ generated by hyponym image $i$, we first adopt a textual projector $g_{text}$ and a visual projector $g_{vis}$ to project $h$ and $p$ to a shared latent space $z$ respectively. During the process, the last dimension of the projected vector is treated as uncertainty and the remaining dimensions as projected representations:
\begin{align}
    \label{eq:projection}
    z_h, u_h = g_{text}(h) = \boldsymbol{W}_{text}h + \boldsymbol{b}_{text},\nonumber\\
    z_p, u_p = g_{vis}(p) = \boldsymbol{W}_{vis}p + \boldsymbol{b}_{vis},
\end{align}
where $\boldsymbol{W}_{text}$, $\boldsymbol{b}_{text}$, $\boldsymbol{W}_{vis}$, and $\boldsymbol{b}_{vis}$ are learnable parameters, $z_h$ and $z_p$ are projected representations, and $u_h$ and $u_p$ represent the uncertainty of the projection process.
Then, we scale the uncertainty $u$ to lie within $(0, 1)$:
\begin{equation}
    \label{eq:sigmoid}
    \tau_h = \sigma(u_h), 
    \tau_p = \sigma(u_p),
\end{equation}
where $\sigma$ is a sigmoid function.
If a projector projects a representation with greater uncertainty, it suggests that the representation is more invariant to displacement, indicating that the representation itself is more general. Hence, we assume that the similarity between a general textual representation and its corresponding prototype is weak.

To capture the similarity controlled by uncertainty, we then propose to use contrastive learning with a variant temperature.
In this approach, we define the projected hypernym representation $z_h$ as an anchor and its corresponding projected visual prototype $z_p$ as the positive sample. We also include other projected prototype representations $z_{neg}^i$ as negative samples ($1 \leq i \leq N-1$), where $N$ is the number of prototypes in the batch. 
To scale each projected representation $z$, we multiply it by its corresponding uncertainty $\tau$. The proposed Hyper-Proto Constraint (HPC) is given by:
\begin{equation}
    \label{eq:hpc}
        \Lagr_{HPC}=-\log \frac{{e^{\tau_h z_h \cdot \tau_p z_p}}}{e^{\tau_h z_h \cdot \tau_p z_p }+\sum_{i=1}^{N-1}{e^{\tau_h z_h \cdot \tau_{neg}^i z_{neg}^i}}}.
\end{equation}

\subsection{Detection Task}
\label{sec: fusion}
The detection task aims to fuse representations from both textual and visual modalities and determine whether the hypernymy relation holds for the given term pairs.

While the intuitive way to fuse textual and visual representations is to add them directly, our preliminary experiments show that when faced with previously unseen term pairs, adding visual features leads the model to predict a hypernymy relation even when it is not applicable. To address this issue, we design a fusion method to generate more informative representations.

To achieve this, we first project a pair of textual term representations $T=(t_{e}, t_{o})$ and its corresponding visual features $V=(p, v)$ using projection functions $g_{text}$ and $g_{vis}$ mentioned in Sec \ref{sec:hpc}, which transform them into $z_{e}$, $z_{o}$, $z_{p}$, and $z_{v}$, respectively. We then calculate the cosine similarity between term representations of the textual and visual modalities using the following equations:
\begin{equation}
    \label{eq:alhpa}
    \alpha_{e} = \frac{1}{1 + e^{-cos(z_{e}, z_{p})}},\;
    \alpha_{o} = \frac{1}{1 + e^{-cos(z_{o}, z_{v})}}.
\end{equation}
We assume that if the projected representations are distinct, indicating that the textual and visual semantics are different, additional visual semantics may provide misleading information.
In the context of user search, an instance of a term pair, such as (\texttt{Zebra}, \texttt{Okapi}), may pose a challenge to the model's performance as \texttt{Okapi}, a species of giraffe that features black and white stripes, shares some visual characteristics with \texttt{Zebra}. However, this sort of confusion can be mitigated by assessing the similarity between the textual and visual representations of the terms in the given pair. A low similarity score extracted from such an analysis can prevent the model from ascribing an erroneous hypernymy relation.

Thus, the similarity between the projected representations provides automatic threshold control over the weighted sum. The term representations $c$ for each term are then fused as follows:
\begin{equation}
    \label{eq:fusion add}
    c_{e} = (1-\alpha_{e}) t_{e} + \alpha_{e} p,\;
    c_{o} = (1-\alpha_{o}) t_{o} + \alpha_{o} v.
\end{equation}
After fusing the representations, we utilize an MLP module to predict whether the potential hypernymy pair exists in the taxonomy. To enhance the relationship between the two representations, we also include the element-wise product between them in the detection module, which is formulated as:
\begin{equation}
    \label{eq:deicsion}
    \begin{aligned}
        \hat{y} = f_{MLP}(c_{e},  c_{o})
        = \boldsymbol{\W}(c_{e} || c_{o} || c_{e} \odot c_{o}) + \boldsymbol{b},
    \end{aligned}
\end{equation}
where $\boldsymbol{\W}$ and $\boldsymbol{b}$ are learnable parameters, $\odot$ is the element-wise product, and $||$ is the concatenation of vectors. The classification objective can be expressed as:
\begin{equation}
    \label{eq:bce}
    \Lagr_{BCE} = -y\log(\hat{y}) - (1 - y)\log(1 - \hat{y}).
\end{equation}

\subsection{Training and Inference}

\textit{Training.}
During training, the objective is the sum of the above four tasks, which is:
\begin{equation}
    \label{eq:overall}
    \Lagr = \Lagr_{text} + \Lagr_{proto} + \Lagr_{HPC} + \Lagr_{BCE}.
\end{equation}

It is worth mentioning that although $\boldsymbol{R_1}$, $\boldsymbol{R_2}$, and $\boldsymbol{R_3}$ selects in-batch negative samples, the detection task does not automatically generate a negative sample. Thus, we propose to select negatvie samples from a set of potential options, which includes the following terms:
1) Children of the anchor node.
2) Siblings of the anchor node.
3) Random items from user click logs that are neither ancestors nor descants of the anchor node.

\noindent\textit{Inference.}
We partition the inference stage into two distinct steps. The first step involves predicting whether the edge connecting a given pair of terms exists within the taxonomy. Here, we directly fuse the generated textual and visual features and subsequently feed them into the classifier for prediction.

Next, we utilize a top-down bootstrapping strategy to expand the predicted hypernymy relationships within the existing taxonomy.
The strategy iterates to traverse the existing taxonomy in level-order.
At iteration $i$, we reach the level $i$, who has a node set $N_i$, and for each node $n \in N_i$, we attach its newly detected hyponyms to $n$ at level $i + 1$. Then, we continue the process until we reach the last level of the taxonomy, and there is no further edge that requires attachment.
\section{Experiments}
This section details our experimental setup and presents the empirical results of our proposed method compared to baseline methods. We also provide a comprehensive analysis of our method to highlight its strengths and weaknesses.

\subsection{Experimental Setup}

\begin{table}[t!]
    \centering
    \setlength\tabcolsep{8.07424pt}
    \begin{tabular}{c|cccc}
    \toprule
        \textbf{Dataset} & \multicolumn{4}{c}{\textbf{Chinese Taxonomy}} \\
        \midrule
        \textbf{Methods} & Accuracy & Precision & Recall & F1 \\
        \midrule
        TaxoExpan & 52.45 & 67.52 & 56.86 & 61.73 \\
        TMN & 54.30 & \underline{70.39} & 52.07 & 59.86 \\
        HyperExpan & 57.15 & 67.41 & 62.59 & 64.91 \\
        PTE & 57.80 & 68.32 & 67.01 & 67.65 \\
        TaxoEnrich & 56.90 & 61.54 & \underline{86.49} & \underline{71.91} \\
        BERT + MLP & \underline{59.00} & 67.84 & 74.57 & 71.05 \\
        \textbf{Ours} & \textbf{67.75} & \textbf{71.23} & \textbf{87.55} & \textbf{78.55} \\
        \midrule
        \midrule
        \textbf{Dataset} & \multicolumn{4}{c}{\textbf{Semeval-2016}} \\
        \midrule
        \textbf{Methods} & Accuracy  & Precision & Recall & F1 \\
        \midrule
        TaxoExpan & 58.72 & 59.63 & \textbf{96.40} & 73.68 \\
        TMN & 71.26 & \underline{76.16} & 74.07 & 75.10 \\
        HyperExpan & 73.51 & 73.02 & 80.19 & 76.44 \\
        PTE & 52.49 & 55.64 & 79.14 & 65.34\\
        TaxoEnrich & \underline{75.46} & 70.80 & \underline{89.81} & \underline{79.18} \\
        BERT + MLP & 74.37 & 73.84 & 82.57 & 77.96 \\
        \textbf{Ours} & \textbf{82.37} & \textbf{79.22} & 87.77 & \textbf{83.28} \\
    \bottomrule
    \end{tabular}
    \caption{Results of Chinese taxonomy dataset and Semeval-2016 dataset. \textbf{Bold} indicates the best and \underline{underlined} indicates second best.}
    \label{table:main}
    \vspace{-1em}
\end{table}


\noindent\textbf{Baselines.}
We compare our method with other state-of-the-art taxonomy expansion approaches. For fair comparisons, we use term embeddings from BERT model in all methods. We run five times for each method and report the average performance. Baselines include: 

\begin{itemize}
    \item \textbf{TaxoExpan} \citep{10.1145/3366423.3380132} encodes positional information with a graph neural network and uses a linear layer to predict hypernymy relations.
    \item \textbf{TMN} \citep{DBLP:conf/aaai/ZhangSZCSM021} leverages auxiliary signals to augment primal task and adopts channel-wise gating mechanism.
    \item \textbf{HyperExpan} \citep{ma-etal-2021-hyperexpan-taxonomy} uses hyperbolic graph neural networks to encode concept embedding and computes matching scores for given pairs.
    \item \textbf{PTE} \citep{9835349} exploits user-generated content by constructing a user-click graph to augment structural information.
    \item \textbf{TaxoEnrich} \citep{jiang2022taxoenrich} generates taxonomy-contextualized embedding and learns horizontal and vertical representation for each term.
    \item \textbf{BERT + MLP} uses a BERT model as the encoder and an MLP module as the classifier to predict whether hypernymy relations holds.

\end{itemize}

\noindent\textbf{Evaluation Metrics.}
In this paper, we report four important metrics in the taxonomy expansion task: accuracy, precision, recall, and F1 score for every method.


\noindent\textbf{Parameter Settings.}
We utilize the ``bert-base-uncased''
and ``bert-base-chinese''
pre-trained models as our textual backbone, and the ``resnet-101''
model as our visual backbone.
For the prototype embedding table, we set the values of $k$ and $e$ to 1024 and 256, respectively. The temperature hyper-parameters for contrastive learning are set to $\tau_{text} = 0.1$ and $\tau_{proto} = 0.1$. We also set $\alpha = 0.999$ and $\epsilon = 0.001$ for EMA. All parameters, except for the prototypes, are optimized using AdamW \citep{https://doi.org/10.48550/arxiv.1412.6980} with a learning rate of 5e-5.

\begin{table}[t!]
    \centering
    \small
    \begin{tabular}{lcccc}
    \toprule
        \textbf{Methods} & Accuracy & Precision & Recall & F1  \\
        \midrule
        \textbf{Ours} & \textbf{67.75} & \textbf{71.23} & \underline{87.55} & \textbf{78.55} \\
        \midrule
        \multicolumn{5}{l}{\textit{Visual Feature}} \\
        \quad w/o image & 61.20 & 68.13 & 76.23 & 71.95 \\
        \midrule
        \multicolumn{5}{l}{\textit{Prototype Learning}} \\
        \quad w/o HPC & 65.80 & 70.34 & 85.25 & 77.08 \\
        \quad w/o prototype & 62.90 & 64.13 & \textbf{88.58} & 74.40 \\
        \midrule
        \multicolumn{5}{l}{\textit{Representation Fusion}} \\
        \quad w/o $\odot$ in Eq.\ref{eq:deicsion} & \underline{66.95} & \underline{70.98} & 86.29 & 77.89 \\
        \quad w/o $\alpha$ in Eq.\ref{eq:fusion add} & 66.50 & 70.64 & 86.14 & 77.62 \\    
        
    \bottomrule
    \end{tabular}
    \caption{Ablation studies on Chinese taxonomy dataset. We remove certain parts from our method. \textbf{Bold} indicate the best and \underline{underlined} indicate second best.}
    \label{table:ablation}
    \vspace{-1em}
\end{table}

\begin{table}[t!]
    \centering
    \small
    \setlength\tabcolsep{8.07424pt}
    \begin{tabular}{c|cccc}
    \toprule
        \midrule
        \textbf{Setting} & Accuracy & Precision & Recall & F1 \\
        \midrule
        zero-shot & 54.85 & 67.61 & 63.45 & 65.47 \\
        1-shot & 60.55 & 69.85 & 77.98 & 73.69 \\
        2-shot & 64.55 & 70.50 & 78.43 & 74.25 \\
        \midrule
        \textbf{Ours} & \textbf{67.75} & \textbf{71.23} & \textbf{87.55} & \textbf{78.55} \\
    \bottomrule
    \end{tabular}
    \caption{Performance of ChatGPT on Chinese taxonomy dataset.}
    \label{table:chatgpt}
    \vspace{-1em}
\end{table}

\subsection{Main Result}

\noindent\textbf{Model Comparison.} Our method is compared with other baseline methods on two datasets. 
Based on the findings in Table \ref{table:main}, it can be concluded that:
1) Our method outperforms all other methods on both datasets because we integrate visual semantics and representation learning in taxonomy expansion.
More specifically, on the Chinese taxonomy dataset, our model achieves a significant improvement over the second-best method by 8.75\% and 6.64\% on accuracy and F1 score, respectively. On the Semeval-2016 dataset, we improve the performance by 6.91\% and 4.1\% on accuracy and F1 score, respectively.
2) Methods that facilitate representation learning (\textit{TaxoEnrich} and \textit{ours}) outperform those that only encode structural information (others), which demonstrates that encoding terms more precisely is the top priority for the taxonomy expansion task.
3) All methods achieve lower performance on our constructed dataset for the taxonomy expansion task, which reflects the challenges associated with this dataset.
Our dataset includes several hard false parent candidates, such as synonyms, prototypical hypernyms, and siblings of the given query term, making it much more difficult to identify the true parents.
4) Despite achieving the third-best recall on the Semeval-2016 dataset, our method has higher precision than the \textit{TaxoEnrich} method by more than 8\%, with a 2.04\% lower recall.
Additionally, the precision of the expanded edges is more important than the number of them.

\noindent\textbf{Domain Taxonomy Expansion.} Our method is evaluated on the Meituan platform using 3 million query-click pairs collected from user logs within a span of 3 months.
The size of real-world product taxonomies increases from 42,714 to 120,872 edges after applying our model.

\subsection{Ablation Studies}

To assess the impact of each component, an ablation analysis is conducted on our method.
Specifically, we remove one component and assess the effectiveness of the remaining combination of components.
Table \ref{table:ablation} presents the results of these experiments.

From the table, we conclude that:
1) eliminating any component or task decreases performance, highlighting their importance.
2) Visual features are the most critical component, as their removal leads to the highest drop in performance compared to other components.
To verify our assumption, prototypes and images are removed to remove the visual features.
The resulting model's performance is then compared to that of our proposed method.
Therefore, removing visual features leads to a significant drop in the accuracy and F1 score by 6.55\% and 6.6\%, respectively.
The result confirms that visual semantics play a crucial role in taxonomy expansion tasks.
3) The proposed Visual Prototype learning and Hyper-Proto Constraint improve overall performance, pointing out that they extract crucial visual semantics information.
Without HPC, accuracy drops by 1.95\%, and F1 score drops by 1.47\%.
However, without visual prototype learning, the model obtains higher recall but lower precision, highlighting the likelihood of encountering the "Prototypical Hypernym Problem." 
4) Without our representation fusion methods, the model's performance slightly drops, demonstrating that the representation fusion methods generate informative representations.




\begin{figure}[t!]
    \centering
    \includegraphics[width=.95\linewidth]{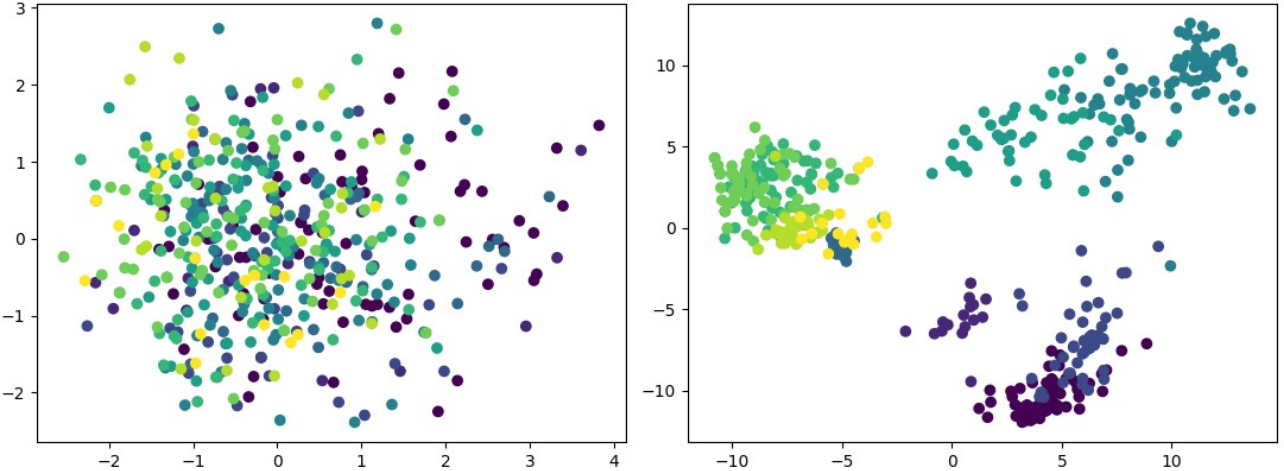}
    \caption{Textual representations before and after applying the Textual Hypernymy Learning task on the training data from the Chinese Taxonomy dataset using PCA.}
    \label{fig:textual}
    \vspace{-1em}
\end{figure}

\subsection{Detailed Analysis}
In this part, we analyze the impact of the three proposed representation learning tasks.

\noindent\textbf{Textual Representation Analysis.}
Figure \ref{fig:textual} illustrates the textual representations before and after training, depicting the textual representation of the Chinese taxonomy dataset's training data using PCA.
We select the top ten hypernyms with the highest number of hyponyms as classes in this visualization.
The figure demonstrates that the Textual Hypernymy Learning task groups semantically similar hyponyms into clusters and distinguishes different hypernyms, which proves the effectiveness of the task.
Notably, some classes are closer to each other because they represent semantically similar hypernyms themselves.

\begin{figure*}[t!]
    \centering
    \includegraphics[width=.9\linewidth]{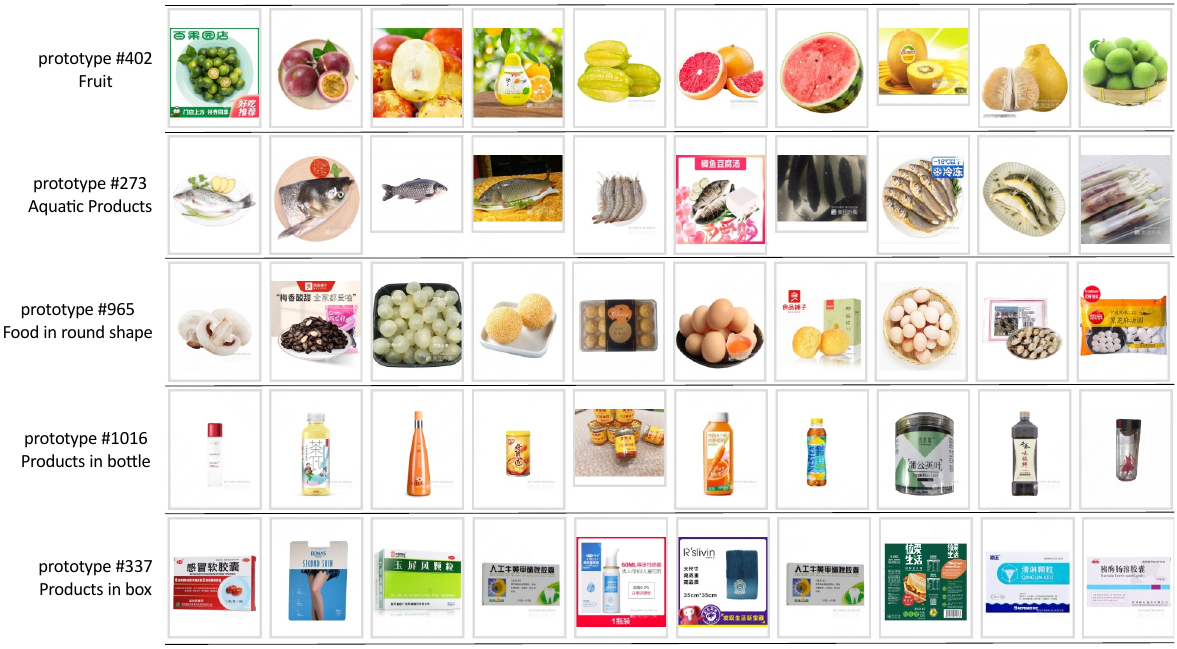}
    \caption{Images from different prototypes. Each line shows images belonging to the cluster of a certain prototype. We provide a brief explanation of the visual semantics of the provided prototype.}
    \label{fig:samples}
    \vspace{-1em}
\end{figure*}

\noindent\textbf{Prototype Analysis.}
To visualize the correspondence between prototypes and images, we select five prototypes from the prototype embedding table and display the images assigned to each prototype cluster in Figure \ref{fig:samples}, where each line corresponds to a distinct prototype.
Some prototypes correspond to textual concepts (e.g. prototype \#402 and \#273), while others represent distinctive visual semantics (e.g. prototype \#965 and \#1016).
For example, prototype \#402 belongs to a fruit subclass, and prototype \#1016 depicts products in bottles that are not represented by textual semantics.
More specifically, given the example input of \texttt{Fruit} and \texttt{Apple Juice} in Figure \ref{fig:intro}, \texttt{Fruit} may correspond to prototype \#402, whereas \texttt{Apple Juice} may correspond to prototype \#1016, facilitating their distinction.
These unique visual semantics present an alternate way to view a term, enhancing the capability to define an unfamiliar term and distinguish a ``Prototypical Hypernym.''

\begin{figure}[t!]
    \centering
    \includegraphics[width=\linewidth]{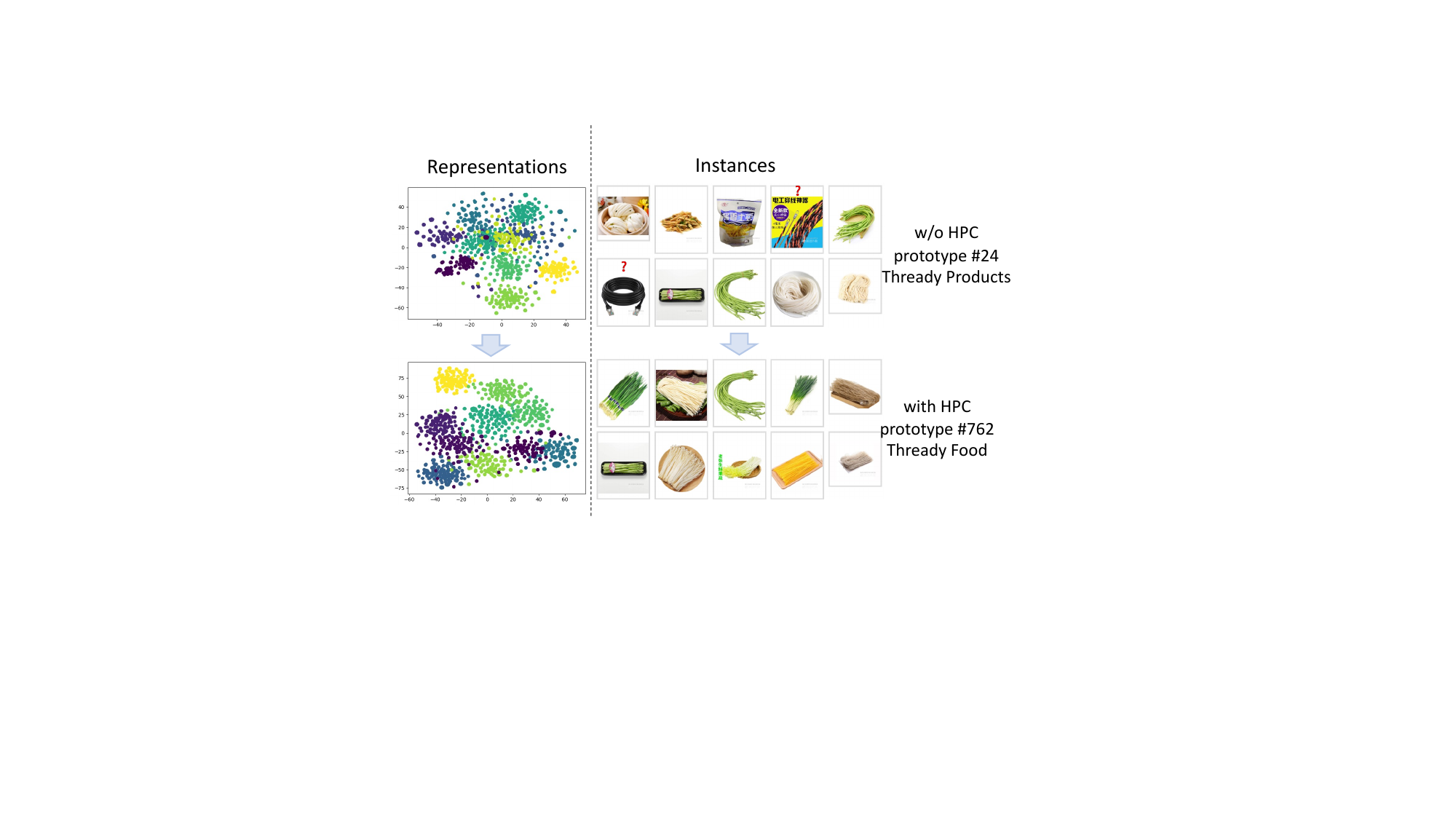}
    \caption{Visual representations and instance images with and without HPC. We present visual representations for training data in Chinese taxonomy dataset using t-SNE and instance images from similar prototypes.}
    \label{fig:visual_all}
    \vspace{-1em}
\end{figure}

\noindent\textbf{Constraint Analysis.}
To demonstrate the effectiveness of the proposed HPC, we visualize the visual representations using t-SNE and select the top ten clusters based on the number of instances.
We also compare prototypes representing similar visual semantics from models with and without HPC.
The visualization is presented in Figure \ref{fig:visual_all}.
The visual representations, portraying more separate clusters, indicate that HPC enhances the model's ability to cluster identical images and separates the clusters further.
From the instances given in the figure, two prototypes representing ``thready objects'' are presented.
However, with HPC, images of \texttt{USB Cable} and \texttt{Rope} are excluded from the cluster, shifting the cluster's semantics from ``thready products'' to ``thready food''.
This suggests that HPC encourages visual prototypes to convey more fine-grained semantics, showing the effectiveness of the HPC.

\noindent\textbf{Comparison with ChatGPT \citep{ouyang2022training}.}
To compare with the current state-of-the-art large language model, we conduct experiments on the Chinese taxonomy dataset, using ``gpt-3.5-turbo''.
The results presented in Table \ref{table:chatgpt} conclude that ChatGPT's mighty power becomes ineffective when compared with our method.
Specifically, despite that multiple samples are given in the prompt, the resulting accuracy drops by 3.2\%, and the F1 score declines by 4.3\%, highlighting the importance of visual semantics.

\subsection{Case Study}


\begin{table*}[t!]
    \centering
    \small
    \begin{tabular}{c|c|c|c|c|c||c|c|c|c|c|c}
    \toprule
        \multicolumn{6}{c||}{\textbf{Corrected Cases}} & \multicolumn{6}{c}{\textbf{Wrong Cases}}\\
        \midrule
        Type & Hyper & Hypo & Truth & Ours & BERT & Type & Hyper & Hypo & Truth & Ours & BERT\\
        \midrule
        UT & \textit{Biscuit} & \textit{Japanese Mochi} & \Checkmark & \Checkmark & {\color{red} \XSolidBrush} & MI & \textit{Fruit Shop} & \textit{Watermelon Cut} & \XSolidBrush & {\color{red}\Checkmark} & \XSolidBrush \\
        UT & \textit{Lunch Box} & \textit{Stainless Steel Snack Cup} & \Checkmark & \Checkmark & {\color{red} \XSolidBrush}  & MI & \textit{Swelling} & \textit{Yunnan Baiyao Ointment} & \XSolidBrush & {\color{red}\Checkmark} & {\color{red}\Checkmark} \\
        PH & \textit{Foam} & \textit{Foaming Gel} & \XSolidBrush & \XSolidBrush & {\color{red}\Checkmark} & LoS & \textit{Kanto Cooking} & \textit{Fish Roe Lucky Bag} & \Checkmark & {\color{red}\XSolidBrush} & \Checkmark \\
        PH & \textit{Loquat} & \textit{Strong Loquat Lotion} & \XSolidBrush & \XSolidBrush & {\color{red}\Checkmark} & LoS & \textit{Baby Products} & \textit{Piggy Page CD} & \Checkmark & {\color{red}\XSolidBrush} & {\color{red}\XSolidBrush} \\
        \bottomrule
    \end{tabular}
    \caption{Case studies of Chinese taxonomy dataset. Corrected cases represent cases that are corrected after introducing visual features. Red marks denote wrong answers. The types indicate the case as either Unseen Term (UT), Prototypical Hypernym (PH), Misleading Image (MI), or Lack of Semantics (LoS).}
    \label{table:case_study}
    \vspace{-1em}
\end{table*}





To provide a more comprehensible view of the efficiency of our assumption, we present the overall performance improvement, and some cases from the Chinese taxonomy dataset in Table \ref{table:case_study} for comparison against the BERT+MLP baseline.

From the statistics of 2,000 test cases, 284 samples have been corrected, whereas only 46 cases were incorrectly predicted, which reaffirms the efficacy of our proposed model in expanding visual taxonomy.

The influence of visual semantics integration can be illustrated further by analyzing some cases in greater detail.
The corrected cases can be divided into two categories:
1) Images provide additional semantics for the model.
For instance, the hyponym \texttt{Japanese Mochi} cannot be linked with the hypernym \texttt{Biscuit} with mere textual semantics. However, by introducing the image of \texttt{Japanese Mochi}, which resembles the other hyponyms of \texttt{Biscuit}, our model effectively predicts \texttt{Japanese Mochi} as a hyponym of \texttt{Biscuit}. 
2) Images differentiate between similar "prototypical hypernyms".
For example, the corrected negative case involves \texttt{Strong Loquat Lotion} that shares similarities in textual semantics with its candidate hypernym \texttt{Loquat}. When visual features are added, the model is still able to differentiate between the two similar terms.

However, it is important to note that incorporating visual semantics into taxonomy expansion may also result in inaccurate predictions, which can be divided into two categories. 
1) Vague images may lead to misleading visual semantics.
For instance, our proposed method may predict \texttt{Watermelon Cut} as a hyponym of a \texttt{Fruit Shop} as a result of the fruit pictures in the hyponyms of a \texttt{Fruit Shop}.
This was observed in our preliminary experiments, and we develop a heuristic fusion method in section \ref{sec: fusion} to address this issue. 
2) There are cases where neither textual nor visual semantics can identify the hypernymy relations, such as the example of \texttt{Piggy Page CD}. In such cases, neither the BERT model nor images can provide detailed semantics.
\section{Conclusions}

This paper introduces visual features into the taxonomy expansion task by proposing a contrastive multitask framework. The framework clusters textual and visual semantics on their respective modalities and bridges the gap between them. Experimental results demonstrate that our proposed framework outperforms state-of-the-art results on numerous datasets. Additionally, visualization analysis confirms that visual features provide diverse semantics.
\section{Acknowledgements}
This work is supported by Shanghai Municipal Science and Technology Major Project (No. 2021SHZDZX0103), Science and Technology Commission of Shanghai Municipality Grant (No. 22511105902), and Shanghai Sailing Program (No. 23YF1409400).

\bibliographystyle{ACM-Reference-Format}
\bibliography{anthology,reference}
\clearpage
\appendix
\section{Datasets Construction}
In this section, we introduce the constructed dataset from Meituan for the taxonomy expansion task.
Moreover, we refine a public dataset, namely the Semeval-2016 task 13, to verify the robustness of our method.

\subsection{Chinese Taxonomy Dataset}
To establish a sizeable product taxonomy dataset comprising images, we rely on the Meituan platform, one of the world's most extensive life service platforms. We will introduce how we build the training, evaluation and test set.

Our training set draw from the existing Meituan product taxonomy, which comprise about 600,000 edges.
However, our observations reveal that 91.67\% of these edges follow the suffix pattern, indicating that the hyponym ends with the hypernym. We define these as suffix hypernymy relations that can be readily resolved through simple string matching. To promote diversity, we eliminate all suffix hypernymy relations and expande our focus on non-suffix hypernymy relations.
We obtain the associated term images using a targeted image search within the Meituan scenario, where the designated term image is the retrieved product image.
With the suffix hypernymy relations eliminated and the images collected, our final training set have 10,697 non-suffix hypernymy pairs, each with a corresponding hyponym image.

For the evaluation and test set, we use user click logs from the Meituan platform, filtered all query-click pairs with the suffix pattern, and ensure that the query is a node within the pre-existing taxonomy. We annotate 2,500 query-click pairs selected at random by five domain experts, where 500 pairs are assigned to the evaluation set.

\subsection{Semeval-2016 Dataset}
We construct the Semeval-2016 dataset to evaluate the stability and effectiveness of our proposed solution, based on Semeval-2016 task 13 \citep{bordea-etal-2016-semeval}. However, this dataset does not contain images for the terms, nor does it have a designated test set. In order to incorporate visual features and establish a test set, we implement a bootstrapping approach. Starting from a randomly selected leaf node from the taxonomy, we take note of its parent node, then remove it from the taxonomy. This process is iterated until 20\% of the nodes have been removed from the taxonomy and deemed suitable for use as the test set. The remaining 80\% of nodes are allocated for the training set.
To enhance the visual representation of each term, we search for each term on Google Images\footnote{https://images.google.com}. The first image returned by the search engine is used as the representative image for each term, given that the search engine has a ranking system that orders images by relevance.
\end{document}